\pdfoutput=1

\documentclass[11pt]{article}

\usepackage[final]{ACL2023}
\usepackage{graphicx}
\usepackage{times}
\usepackage{latexsym}
\usepackage{multirow}
\usepackage{makecell} 
\usepackage{booktabs} 
\usepackage{hyperref} 
\usepackage{float}    
\usepackage[T1]{fontenc}

\usepackage[utf8]{inputenc}

\usepackage{microtype}

\usepackage{inconsolata}

\usepackage{xcolor}       
\usepackage{listings}     
\usepackage{hyperref}     
\usepackage{graphicx}     
\usepackage{fontawesome}  
\usepackage{url}          

\title{CMHG: A Dataset and Benchmark for Headline Generation of Minority Languages in China}

\author{%
      Guixian Xu\textsuperscript{1,2 $\dagger$}  \
      Zeli Su\textsuperscript{1,2}   \
      Ziyin Zhang\textsuperscript{3}  \
      Jianing Liu\textsuperscript{2} \AND
      XU Han\textsuperscript{1,2} \
      Ting Zhang\textsuperscript{1,2} \
      Yushuang Dong\textsuperscript{1,2}\
      \vspace{6pt}\\
      \textsuperscript{1}Key Laboratory of Ethnic Language Intelligent Analysis and Security Governance of MOE \\
      \textsuperscript{2}Minzu University of China \
      \textsuperscript{3}Shanghai Jiao Tong University \\
      \texttt{\{rickamorty,guixian\_xu,hanxu,jianing\_liu,yushuangdong\}@muc.edu.cn}\\
      \texttt{daenerystargaryen@sjtu.edu.cn} \
      \texttt{tozhangting@126.com} \textsuperscript{$\dagger$} Corresponding author
}

\begin{document}
\maketitle
\begin{abstract}
Minority languages in China, such as Tibetan, Uyghur, and Traditional Mongolian, face significant challenges due to their unique writing systems, which differ from international standards. This discrepancy has led to a severe lack of relevant corpora, particularly for supervised tasks like headline generation. To address this gap, we introduce a novel dataset, Chinese Minority Headline Generation (CMHG), which includes 100,000 entries for Tibetan, and 50,000 entries each for Uyghur and Mongolian, specifically curated for headline generation tasks. Additionally, we propose a high-quality test set annotated by native speakers, designed to serve as a benchmark for future research in this domain. We hope this dataset will become a valuable resource for advancing headline generation in Chinese minority languages and contribute to the development of related benchmarks.

\hspace{-10pt}\includegraphics[width=1em,height=1em]{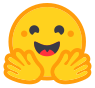}\hspace{-3pt} {\small~\url{https://huggingface.co/KEVVVV/CMHG}}

\end{abstract}

\section{Introduction}

Recently, the rapid development of large language models (LLMs) has been fueled by the availability of high-quality pre-training data. However, these advancements have primarily benefitted high-resource languages such as English and Chinese. In contrast, many languages with substantial user bases remain excluded due to the scarcity of suitable corpora, especially for specific tasks like \textit{headline generation}. This exclusion poses challenges for both academic research and the practical application of AI technologies.

\begin{figure}
    \centering
    \includegraphics[width=1\linewidth]{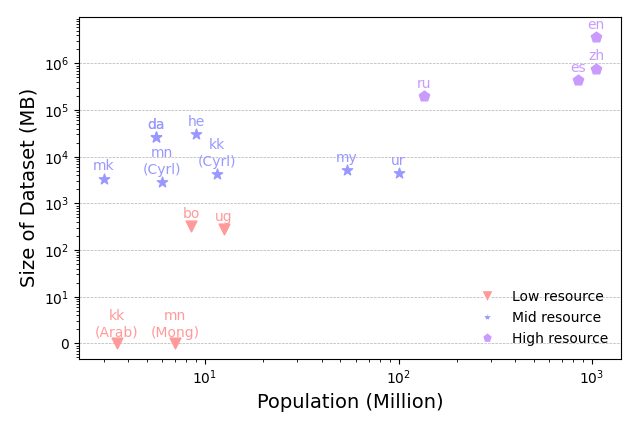}
    \caption{The relationship between population size and dataset size in OSCAR (y-axis, in MB) for various high-, middle-, and low-resource languages.}
    \label{fig-1}
\end{figure}

This paper focuses on underrepresented minority languages in China, including Tibetan, Uyghur, and Mongolian, which have rich linguistic and cultural significance but suffer from a lack of resources. Although these languages appear in multilingual datasets like OSCAR~\cite{OSCAR} and CulturaX~\cite{culturax}, quality issues limit their usefulness. As shown in Figure~\ref{fig-1}, there is a clear gap between the large speaker populations of these languages and the small amount of data available in major corpora. Studies also reveal problems with data quality: \citet{mc2} found that 34\% of the Uyghur data in CulturaX contains Kazakh or Arabic texts, pointing to issues like language misidentification and noise. These challenges of data scarcity and quality undermine efforts to build effective natural language processing (NLP) systems for these communities.

Moreover, there is a complete lack of open-source datasets tailored for headline generation in these minority languages. This gap hinders the development of supervised methods and benchmarks for headline generation tasks. To address this limitation, we introduce \textbf{Chinese Minority Headline Generation (CMHG)}, a novel dataset specifically designed for headline generation in Tibetan, Uyghur, and Mongolian. CMHG consists of \textbf{100,000 Tibetan samples} and \textbf{50,000 samples each} for Uyghur and Mongolian. In addition to the main dataset, we collaborated with native speakers of these languages to further ensure data quality. From the existing data, we selected \textbf{3,000 samples} for each language and conducted a detailed annotation process to evaluate the alignment and quality of the headlines. These samples were reviewed by multiple annotators for each language, and only the data deemed high-quality by consensus among native annotators was retained. This subset of data provides a reliable \textit{benchmark} for future research, ensuring consistency and reproducibility in evaluating headline generation models. By combining a large-scale dataset with carefully curated benchmark samples, CMHG bridges the gap in resources for headline generation in Chinese minority languages.

In summary, this paper makes the following key contributions:
\begin{itemize}
    \item We present \textbf{CMHG}, a novel and large-scale open-source dataset specifically designed for headline generation in three Chinese minority languages: Tibetan, Uyghur, and Mongolian. We release this dataset.
    \item We provide a carefully curated benchmark test set, annotated by native speakers, to ensure high-quality evaluation and support transparent, reproducible research in headline generation for Chinese minority languages.
\end{itemize}

By introducing CMHG, we aim to fill the resource gap and pave the way for advancing natural language processing research on underrepresented languages.

\section{Data Sources}

The Chinese Minority Headline Generation (CMHG) dataset, proposed in this paper, is sourced from various online platforms in China, including government documents and news articles (detailed list in Appendix~\ref{appendix-A}). We used web crawlers to collect the data, where the webpage title serves as the headline and the main text as the source content. To ensure data quality and reliability, we applied a thorough cleaning process, with the main methods outlined as follows:

\begin{itemize}
    \item \textbf{Removal of Non-Textual Content}: We filtered out non-textual elements such as advertisements, pop-ups, navigation bars, and multimedia (e.g., images, videos, and audio files). This ensured that only relevant text was retained.
    \item \textbf{Duplicate Detection and Removal}: We identified and removed duplicate entries to avoid redundancy in the dataset, which could potentially bias the headline generation models.
    \item \textbf{Text Normalization}: We converted all characters to uniform encoding and removing any extraneous white spaces, special characters, or formatting inconsistencies.
    \item \textbf{Language Purity Check}:We sourced content from monolingual websites and used regex-based filtering to ensure linguistic purity in manually annotated data. Controlled noise was allowed in training data to improve model robustness. Additionally, we evaluated several existing Language Identification tools that claimed to support our target languages but found high false-positive rates.
\end{itemize}

These steps were essential to ensure that the final dataset consisted of high-quality, relevant, and clean text that could be reliably used for our tasks.

\section{Annotation}

After crawling the data, we further enhanced the quality of the three languages' evaluation set by native speaker annotation. For each language, two annotators assessed  the quality of the title-content matching. A total of 3000 randomly selected samples from the crawled data were annotated, with the task focused on evaluating how well the article titles matched the content of the articles. 

\subsection{Annotation Guidelines}

The annotation process was designed to ensure the reliability and consistency of the evaluations. The annotators were provided with the following specific guidelines:


\begin{table*}[h!]
    \centering
    \renewcommand{\arraystretch}{1.2} 
    \setlength{\tabcolsep}{5pt} 
    \begin{tabular}{c c c c c c c}
        \toprule 
        \textbf{Language} & \textbf{Size} & 
        \multicolumn{1}{c}{\textbf{\parbox[c]{3.5cm}{\centering Length In Token\\(Title/Content)}}} & 
        \multicolumn{1}{c}{\textbf{\parbox[c]{3.5cm}{\centering Length In Characters\\(Title/Content)}}} &
        \textbf{Cohen's $\kappa$} & \textbf{ICC} & 
        \multicolumn{1}{c}{\textbf{\parbox[c]{1.5cm}{\centering Same\\Tendency}}} \\
        \midrule 
        Tibetan  & 2901 & 12.3 / 376.7 & 74.0 / 1884.1 & 0.71 & 0.80 & 1.00 \\
        Mongolian& 2931 & 27.2 / 429.8 & 136.1 / 2149.0 & 0.28 & 0.42 & 0.85 \\ 
        Uyghur   & 2950 & 30.2 / 815.7 & 151.0 / 4078.5 & 0.44 & 0.67 & 1.00 \\
        \bottomrule 
    \end{tabular}
    \caption{Annotation Results and Inter-Annotator Agreement Analysis for Valid Samples in the CMHG Dataset. Token length is counted by the CINO~\citep{cino} tokenizer; character length by raw character count.}
    \label{tab:annotation_results}
\end{table*}

\begin{itemize}
    \item \textbf{Task Objective:} Annotators were asked to assess the degree of relevance between the title and the content of the article and assign a score accordingly. 
    \item \textbf{Title Evaluation:} Annotators were first asked to identify any issues between the title and the article content, such as:
        \begin{itemize}
            \item \textit{Incomplete Article:} whether the article content is incomplete, making it impossible for the title and content to match.
            \item \textit{Text Quality:} whether the title contains spelling, grammatical, or contextual errors that would significantly hinder its match with the article content.
            \item \textit{Other Issues:} whether there is any other noticeable discrepancies between the title and content, such as irrelevance.
        \end{itemize}
    If no major issues were identified, the title-content match was considered "Normal."
    
    \item \textbf{Matching Score:} Annotators were instructed to rate the match based on how well the title corresponded with the article content. The scoring system was as follows:
        \begin{itemize}
            \item \textbf{1 point:} Completely Mismatched (The title is entirely unrelated to the content).
            \item \textbf{2 points:} Slightly Mismatched (The title is related to the content but does not align with the main theme).
            \item \textbf{3 points:} Slightly Inaccurate (There is some connection, but it is not fully aligned).
            \item \textbf{4 points:} Uncertain (The relationship between title and content is unclear or ambiguous).
            \item \textbf{5 points:} Slightly Matched (There is a strong connection, but there are some inconsistencies).
            \item \textbf{6 points:} Well Matched (The title matches the content with only minor discrepancies).
            \item \textbf{7 points:} Fully Matched (The title perfectly corresponds to the content).
        \end{itemize}
\end{itemize}

\subsection{Consistency and Quality Control}

To ensure consistency and accuracy across annotations, multiple annotators evaluated each article. The following steps were implemented to guarantee the quality of the annotations:

\begin{itemize}
    \item \textbf{Consistency Check:} An annotation was considered invalid if the score differed by more than 2 points from the majority of annotators. Additionally, if an annotator’s judgment deviated significantly from the majority opinion (e.g., the majority rated the title as "matching," but the annotator rated it as "not matching"), the annotation would be discarded.
    \item \textbf{Handling Invalid Annotations:} Invalid annotations were removed, and the annotators are incentivized to not produce such annotations.
\end{itemize}

\subsection{Incentive System}

To encourage careful and consistent annotation work, we implemented a reward-based incentive system:

\begin{itemize}
    \item Scores \textbf{< 4} or \textbf{$\geq$ 4} are considered as different tendencies: 
        \begin{itemize}
            \item Scores \textbf{< 4} indicate a \textit{non-aligned} tendency.
            \item Scores \textbf{$\geq$ 4} indicate an \textit{aligned} tendency.
        \end{itemize}
    \item Annotation whose tendency aligns with the majority will receive \textbf{0.25 RMB}.
    \item Annotation that aligns with the majority tendency and furthermore deviates by no more than 1.5 points from the average score will receive an additional \textbf{0.25 RMB}.
\end{itemize}

Annotators were strongly encouraged to adhere to the guidelines to ensure the high quality and consistency of the dataset annotations.

\subsection{Annotation Results}

In general, the data we retained showed a high degree of quality. After removing the data flagged as mismatched or erroneous by the annotators, we retained the samples with an average score above 4. The final number of valid samples and the average length for each language are shown in Table~\ref{tab:annotation_results}. Most retained samples scored 7, achieving an average score of 6.9/7, which confirms the effectiveness of our native-speaker-guided annotation process.

The average title and content lengths reflect the linguistic characteristics of these minority languages. Tibetan titles and contents have shorter average lengths (12.3 tokens/74.0 characters for titles, 376.7 tokens/1884.1 characters for contents) compared to the significantly longer Mongolian and Uyghur samples.

\subsection{Inter-Annotator Agreement Analysis}
We assessed inter-annotation agreement using Cohen’s $\kappa$, ICC, and Same Tendency Rating for each language (Table~\ref{tab:annotation_results}). Despite some variability in specific ratings and a low Cohen's kappa for B0 and MN groups, the Same Tendency Rating was consistently high, indicating agreement on trends and supporting data reliability.


\section{Experiment}

In this section, we evaluate some of the most popular models available for Tibetan, Mongolian, and Uyghur on CMHG, including finetuning small encoder-decoder models and few-shot evaluation of LLMs.


\subsection{Experimental Settings}

\textbf{Fine-tuned Models:}  
The small models, \textbf{cino-cum} (\textit{which uses the \textbf{cino}~\cite{cino} encoder, based on the XLM-R model tailored for Chinese minority languages, and a transformer decoder in a seq2seq architecture}) and \textbf{swcm}~\cite{swcm} (\textit{which is based on the same structure as cino-cum, but incorporates shared weight optimization across the encoder and decoder for improved performance across languages}), are fine-tuned on non-annotated data from the CMHG dataset. These models are then evaluated using high-quality annotated data to assess their headline generation performance. The fine-tuning is conducted on raw, non-annotated data, while the evaluation is done using a set of annotated samples to measure the ROUGE-L scores.

\textbf{Few-shot Models:}  
The large models, \textbf{Qwen2.5-72B}~\cite{qwen2} and \textbf{LLaMA3.1-70B}~\cite{llama3} use a 2-shot learning paradigm, where two annotated samples are dynamically inserted as examples within the input of each  annotated sample.

Detailed training configurations and hyperparameters are provided in Appendix~\ref{appendix-B}.

\subsection{High-Quality Small Sample Experiment}

Given that evaluating large models like \textbf{Qwen2.5-72B} and \textbf{LLaMA3.1-70b} with nearly 3,000 annotated samples per language is resource-intensive, we also selected a high-quality subset for evaluation to facilitate future works. Specifically, we chose the top 500 annotated samples based on evaluation scores to create a high-quality small sample version, enabling more efficient performance assessment while maintaining data quality.

\begin{table}[ht]
  \centering
  \begin{tabular}{c c c c c}
    \hline
    \textbf{Model} & \textbf{Size} & \textbf{bo} & \textbf{mn} & \textbf{ug} \\[0.5ex]
    \hline
    \textbf{cino-cum} & 411M & 0.20 & 0.12 & 0.09 \\
    \textbf{swcm} & 457M & 0.23 & 0.18 & 0.15 \\
    \textbf{Qwen2.5} & 72B & 0.24 & 0.32 & 0.29 \\
    \textbf{LLaMA3.1} & 70B & 0.34 & 0.30 & 0.35 \\
    \hline
  \end{tabular}
  \caption{Model Parameters and ROUGE-L F1 Scores across all annotated data}
  \label{tab:all_results}
\end{table}

\begin{table}[ht]
  \centering
  \begin{tabular}{c ccc}
    \hline
    \textbf{Model} & \textbf{bo} & \textbf{mn} & \textbf{ug} \\[0.5ex]
    \hline
    \textbf{cino-cum} & 0.21 & 0.13 & 0.10 \\
    \textbf{swcm} & 0.23 & 0.17 & 0.14 \\
    \textbf{Qwen2.5} & 0.24 & 0.29 & 0.34 \\
    \textbf{LLaMA3.1} & 0.34 & 0.31 & 0.34 \\
    \hline
  \end{tabular}
  \caption{ROUGE-L F1 Score in High-Quality data}
  \label{tab:simplified_results}
\end{table}

\subsection{Results and Discussion}

The experimental results are summarized in Table~\ref{tab:all_results}, which presents the performance of the models on the CMHG dataset across the three languages: Tibetan (bo), Uyghur (ug), and Mongolian (mn). The fine-tuning results for the small models, \textbf{cino-cum} and \textbf{swcm}, show that both models achieved competitive ROUGE-L scores, demonstrating that fine-tuning with the CMHG dataset enables the models to generate concise and contextually accurate headlines for all three languages. This indicates that the large amount of non-annotated data collected in the CMHG dataset plays a crucial role in enhancing model performance for these underrepresented languages.

For the large models, \textbf{Qwen2.5-72B} and \textbf{LLaMA3.1-70B}, the few-shot results, as shown in Table~\ref{tab:all_results} and Table~\ref{tab:simplified_results}, reveal strong performance across both small and large sample tests. This demonstrates that the models exhibit high-quality headline generation capabilities, regardless of the sample size, highlighting the effectiveness of using small annotated datasets for evaluating model performance. The ability of these models to perform well with just a few annotated samples supports the idea that the CMHG dataset, with its carefully curated annotated samples, can serve as a reliable benchmark for future research and evaluation in headline generation for minority languages.

Overall, both the fine-tuning and few-shot learning approaches contribute significantly to advancing headline generation for minority languages, and the CMHG dataset proves to be a valuable resource for further research in this area.

\section *{Limitations}
Despite CMHG's significant contribution to headline generation for Chinese minority languages, some limitations remain. First, while the CMHG dataset represents a substantial effort to address the data scarcity issue for Tibetan, Uyghur, and Mongolian, the availability of high-quality linguistic resources for these languages is still limited compared to high-resource languages. The scarcity of large-scale annotated datasets for other minority languages in China and beyond further highlights the need for continued efforts to expand the scope of language resources. Additionally, the current dataset focuses primarily on headline generation tasks, leaving other NLP applications underexplored. Future work will aim to broaden the dataset to include more minority languages and diverse NLP tasks, alongside collaborations with native speakers and linguistic experts to enhance data quality and coverage, fostering more inclusive and comprehensive NLP research for underrepresented languages.
\section *{Ethical Statements}
All artifacts in this study are intended for research purposes only, and copyright (where applicable)remains with the original authors or publishers

\section*{Acknowledgements}
This research was supported by the Joint Research Project of Li'an International Education Innovation Pilot Zone, Hainan Province, China (Grant No: 624LALH006).

\bibliography{reference}
\bibliographystyle{acl_natbib}

\appendix

\clearpage
\appendix

\section{ Dataset Details} \label{appendix-A}

\subsection{1. Dataset Size and Domain Distribution}

Table~\ref{tab:dataset_size} provides the size statistics and domain composition of the CMHG dataset for each language.

\section{Training Details} \label{appendix-B}

\subsection*{Fine-tune Training Details}

\phantom{.}\vspace{-14pt}

\textbf{Hardware:} NVIDIA A5000 GPU, 24 GB RAM, Intel i7 CPU.

\textbf{Software:} Ubuntu 20.04, CUDA 11.7, PyTorch 2.3

\subsubsection*{Training Configurations} \phantom{.}\vspace{-14pt}

\textbf{Local Batch Size:} 20

\textbf{Gradient Accumulation Steps:} 4

\textbf{Global Batch Size:} 80

\textbf{Epochs:} 50

\textbf{Optimizer:} AdamW with $\beta_1 = 0.9$, $\beta_2 = 0.999$

\textbf{Learning Rate:} 1e-4

\textbf{Warm-up:} Linear warm-up for the first epoch, gradually increasing the learning rate from 1e-5 to 1e-4.

\subsection*{Few-shot Training Details}

\phantom{.}\vspace{-14pt}

In the few-shot setting, the model is provided with a prompt and a few examples to generate a task-specific output. For this task, the prompt is designed to help the model generate concise and accurate headline in Tibetan, Uyghur, or Mongolian based on a provided passage and its title. The examples are structured to guide the model’s behavior in generating the expected output.

\onecolumn    

\subsubsection*{Prompt} \phantom{.}\vspace{-14pt}

Based on the provided passage with title and content, generate a concise and accurate headline in \textbf{Tibetan/Uyghur/Mongolian}:
\begin{quote} \textbf{Example 1/2}: \\ Content: \{Passage\} \\ Title: \{Title of the passage\}

\textbf {Example 2/2}: \\ Content: \{Passage\} \\ Title: \{Title of the passage\}

\textbf {Task:} \\ Content: \{Passage\} \\ Title: \end{quote}

\begin{table*}[h!]
\centering
\begin{tabular}{lcccc}
\toprule
\textbf{Language} & \textbf{Annotated} & \textbf{Raw} & \textbf{Gov. Docs (\%)} & \textbf{News (\%)} \\
\midrule
Tibetan (bo) & 2,901 & 100,000 & 66 & 34 \\
Mongolian (mn) & 2,931 & 50,000 & 100 & - \\
Uyghur (ug) & 2,950 & 50,000 & 85 & 15 \\
\bottomrule
\end{tabular}
\caption{Dataset size and domain distribution for each language in CMHG. The Mongolian news percentage is zero due to limited availability of news media resources in this language.We use the structural consistency of government websites and news sources to minimize noise during the data filtering process.}
\label{tab:dataset_size}
\end{table*}

\subsection{2. List of Crawled Websites}  

Table~\ref{tab:crawled_websites} lists the websites and URLs used for data crawling.

\begin{table*}[h!]
\centering
\begin{tabular}{lp{6cm}c}
\toprule
\textbf{Website Name} & \textbf{URL} & \textbf{Language} \\
\midrule
Qinghai Lake Website (Tibetan Version) & \url{https://www.amdotibet.cn} & BO \\
China Tibet News Network & \url{https://tb.xzxw.com} & BO \\
Bon Religion Website & \url{http://www.himalayabon.com} & BO \\
Kamba Satellite TV Network & \url{http://tb.kangbatv.com} & BO \\
Qinghai Tibetan Language Radio and TV Station& \url{http://www.qhtb.cn} & BO \\
China Tibetan Calligraphy Website & \url{http://www.zgzzsfw.com} & BO \\
Inner Mongolia Government Website & \url{https://mgl.nmg.gov.cn} & MN \\
Hulunbuir City Government Website & \url{http://mgl.hlbe.gov.cn} & MN \\
Xilingol League Government Website & \url{http://mgl.zlq.gov.cn} & MN \\
Ula'gae Government Website & \url{http://mgl.wlgglq.gov.cn} & MN \\
Chifeng City Government Website & \url{http://mgl.chifeng.gov.cn} & MN \\
Tongliao City Government Website & \url{http://mgl.tongliao.gov.cn} & MN \\
Aksu News Network & \url{https://uy.aksxw.com} & UG \\
Nur Network & \url{https://www.nur.cn} & UG \\
Tianshan Net & \url{http://uy.ts.cn} & UG \\
Xinjiang Government Website & \url{https://uygur.xinjiang.gov.cn} & UG \\
Xinjiang Daily Website & \url{http://xjrbuy.ts.cn} & UG \\
\bottomrule
\end{tabular}
\caption{List of websites used for data crawling.}
\label{tab:crawled_websites}
\end{table*}

\end{document}